\documentclass[10pt,twocolumn,letterpaper]{article}

\usepackage[pagenumbers]{cvpr} % To force page numbers, e.g. for an arXiv version

\newcommand{\lpips}{\scalebox{0.8}{LPIPS$\downarrow$}}

\newcommand{\ssim}{\scalebox{0.8}{SSIM$\uparrow$}}
\newcommand{\psnr}{\scalebox{0.8}{PSNR$\uparrow$}}

\usepackage{soul}
\definecolor{lightblue}{rgb}{.8,.95,1}
\sethlcolor{lightblue}
\newcommand{\hb}[1]{\hl{\textbf{#1}}}

\definecolor{myred}{rgb}{0.99, 0.82, 0.83} 
\definecolor{myyellow}{rgb}{0.98, 0.97, 0.93} 
\definecolor{mygreen}{rgb}{0.94, 0.99, 0.94}

\definecolor{cvprblue}{rgb}{0.21,0.49,0.74}
\usepackage[pagebackref,breaklinks,colorlinks,allcolors=cvprblue]{hyperref}
\usepackage[ruled,vlined]{algorithm2e}
\usepackage{multirow}
\usepackage{stfloats}
\usepackage{caption}

\title{HybridGS: Decoupling Transients and Statics \\ with 2D and 3D Gaussian Splatting}

\author{
    Jingyu Lin$^{1}$ \quad
    Jiaqi Gu$^{2}$ \quad
    Lubin Fan$^{2}$ \quad
    Bojian Wu$^{3}$ \quad
    Yujing Lou$^{4}$ \quad \\
    Renjie Chen$^{1}$ \quad 
    Ligang Liu$^{1}$ \quad
    Jieping Ye$^{2}$
    \\
        \textsuperscript{1}University of Science and Technology of China
        \quad
        \textsuperscript{2}Individual Researcher
        \quad
        \\
        \textsuperscript{3}Zhejiang University
        \quad
        \textsuperscript{3}Shanghai Jiaotong University
    \\
    \textbf{\url{https://gujiaqivadin.github.io/hybridgs/}}
}

\begin{document}

\twocolumn[{%
\renewcommand\twocolumn[1][]{#1}%
\maketitle
\vspace{-3em}
\begin{center}
    \includegraphics[width=1\textwidth]{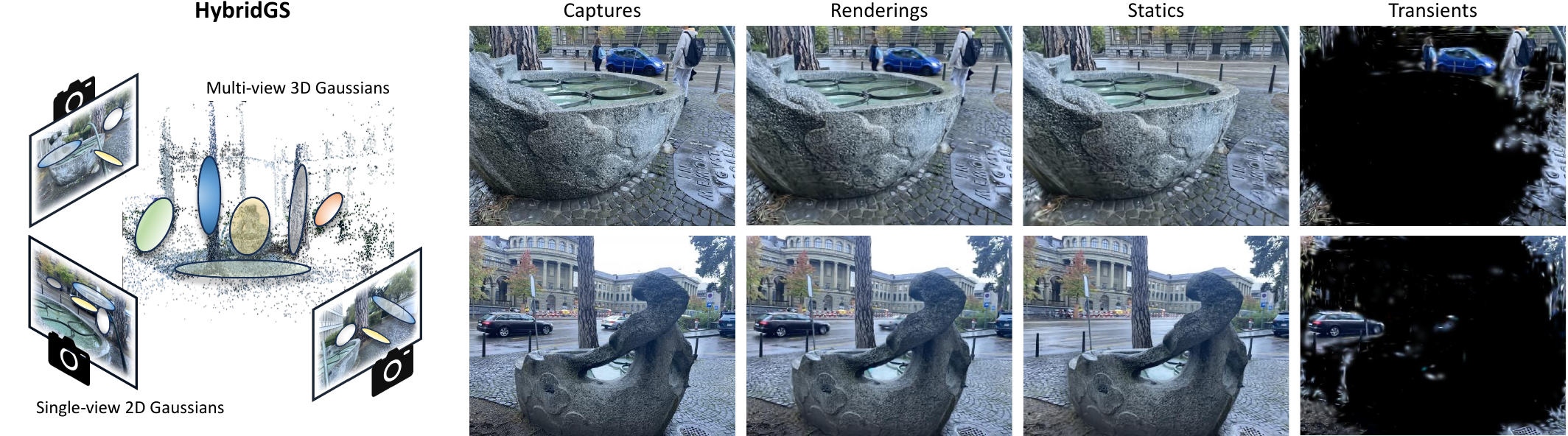}
    \vspace{-1.5em}
    \captionof{figure}{\textbf{HybridGS} is the first hybrid representation that combines multi-view consistent 3D Gaussians and single-view independent 2D Gaussians, which is used to decouple the transients and statics presented in the scene. Our results demonstrate reasonable decompositions.\vspace{1em}}
    \label{fig:teaser}
\end{center}
}]
\begin{abstract}
Generating high-quality novel view renderings of 3D Gaussian Splatting (3DGS) in scenes featuring transient objects is challenging. We propose a novel hybrid representation, termed as \textbf{HybridGS}, using 2D Gaussians for transient objects per image and maintaining traditional 3D Gaussians for the whole static scenes. Note that, the 3DGS itself is better suited for modeling static scenes that assume multi-view consistency, but the transient objects appear occasionally and do not adhere to the assumption, thus we model them as \textbf{planar objects from a single view}, represented with 2D Gaussians. Our novel representation decomposes the scene from the perspective of fundamental viewpoint consistency, making it more reasonable. Additionally, we present a novel multi-view regulated supervision method for 3DGS that leverages information from co-visible regions, further enhancing the distinctions between the transients and statics. Then, we propose a straightforward yet effective multi-stage training strategy to ensure robust training and high-quality view synthesis across various settings. Experiments on benchmark datasets show our state-of-the-art performance of novel view synthesis in both indoor and outdoor scenes, even in the presence of distracting elements.
\end{abstract} 
\section{Introduction}
\label{sec:intro}

3D Gaussians Splatting (3DGS)~\cite{Kerbl20233DGS} has recently gained popularity for novel view synthesis due to its high-quality rendering, efficiency and low-memory cost. Its applications span across virtual reality, augmented reality, and robotics, etc~\cite{wang2024dc, roessle2024l3dg, Charatan2023PixelSplat3G, shi2024language, Liang2023GSIR3G, Wu_2024_CVPR, Chen2023Textto3DUG, li2024art3d3dgaussiansplatting, zhou2024drivinggaussian}. Typically, most current approaches assume that the input images are posed and free of noise, which are the prerequisite needed to represent a complete scene well. However, this assumption is often not met. For example, images taken casually with a mobile phone usually contain messy dynamic objects, or transient objects. Therefore, it is hard to model the scene because there are always many transient occlusions. Technically, 3DGS is capable of modeling static scenes effectively because the static objects in images satisfy geometric consistency constraints across different viewpoints. However, transient objects do not follow this assumption. If the images containing transient objects are naively used in training 3DGS, such variations can lead to difficulties in achieving coherent blending of individual Gaussian representations, potentially resulting in artifacts or inaccuracies.

To address this issue, numerous methods have been proposed to improve the scene representation ability involving transient objects. In the era of Neural Radiance Fields (NeRF)~\cite{mildenhall2020nerf}, RobustNeRF~\cite{Sabour2023RobustNeRFID} introduced a robust loss function that minimizes the influence of photometrically inconsistent observations during training, thereby enhancing the overall quality of the representation. NeRF On-the-go~\cite{Ren2024NeRF} was the first to leverage the uncertainty of transients by incorporating DINOv2~\cite{oquab2024dinov} features into a shallow MLP. It focused on loss supervision in the areas of low uncertainty to effectively remove transients from dynamic scenes. These two methods have established widely-used benchmark datasets featuring transients, aiming to evaluate the reconstruction of clear static scenes with improved results of novel view synthesis. For 3DGS, SpotLessSplats (SLS-mlp)~\cite{sabourgoli2024spotlesssplats} integrated the designs of those aforementioned methods, employing adaptive method to detect outliers through clustering in the feature space, achieving the best current results.

However, these previous approaches neglect two critical issues:
(1) \textbf{The ambiguity between semantics and transients}. Utilizing features from visual models to capture the semantic characteristics of transients can inherently lead to confusion in distinguishing between them.
(2) \textbf{The absence of explicit transient modeling}. These methods primarily focus on achieving a robust statics by merely mitigating the impact of transients, rather than comprehensively modeling the whole scene themselves. 

In response to the issues mentioned above, our key observation is that  transients lack multi-view consistency and usually only appear in a single view. Therefore, we consider them as \textbf{planar objects at that view}. That is to say, this planar representation is a unique feature of the current view. 
Based on this, we propose \textit{\textbf{HybridGS}}, a hybrid representation with 2D Gaussians for transient objects per image and 3D Gaussians for the whole static scenes as illustrated in Fig.~\ref{fig:teaser}, which naturally decouples the transients and statics based on their intrinsic properties.
With this novel representation, while modeling transients as planar objects, the fundamental consistency of viewpoints is also maintained for statics.
To further enhance the performance, a multi-view regulated supervision method for 3DGS is adopted to incorporates information from co-visible regions across multiple views, leading to more accurate representation.
Additionally, we employ a simple yet effective multi-stage training strategy that ensures robust training and consistent view synthesis across various scenarios, thereby guaranteeing better stability and convergence during the training process.
Extensive experiments on benchmark datasets demonstrate our state-of-the-art quality of novel view synthesis in both indoor and outdoor scenes, leading to the accurate decoupling of transients and statics as well as high efficiency in training and inference stage.
The contributions are summarized as follows.

\begin{itemize}
    \item We are the first to introduce a novel hybrid  representation that combines image-specific 2D Gaussians with static 3D Gaussians, enabling effective modeling of transient objects within casually captured images.
    \item We develop a multi-view supervision scheme for 3DGS that utilizes overlapping regions across multiple views, which enhances the model's capability to distinguish between static and transient elements, ultimately improving the overall quality of the novel view synthesis.
    \item Our HybridGS achieves the state-of-the-art performance in benchmark datasets, which outperforms the previous methods and sets a new standard for novel view synthesis in the scenes with transients.
\end{itemize}
\section{Related Work}
\label{sec:related}

\subsection{Novel View Synthesis}
NeRF (Neural Radiance Fields)~\cite{mildenhall2020nerf, barron2021mipnerf, barron2022mipnerf360, Kulhnek2023TetraNeRFRN, mueller2022instant, rematas2022urf, Tancik2022BlockNeRFSL, Tancik2023NerfstudioAM, Lin2024VastGaussianV3} revolutionizes 3D scene reconstruction by representing scenes as continuous neural networks that map 3D coordinates to color and density values, which allows for the synthesis of photorealistic views from arbitrary camera positions.
Recently, 3D Gaussian Splatting (3DGS)~\cite{Kerbl20233DGS, hierarchicalgaussians24, Yu2024MipSplatting, Yu2024GOF, scaffoldgs, taming3dgs, Huang2DGS2024} has gained attention as an effective method for novel view synthesis. It starts with an initial point cloud and transforms it into optimizable 3D Gaussian primitives, known for their high-quality rendering and fast processing speeds. Nevertheless, existing methods require the assumption that only static scenes are present in the input images to effectively utilize multi-view geometric consistency to train a coherent scene representation.

\subsection{Modeling Transients and Statics in NeRF/3DGS}

In casually captured footage of real-world scenes, there are often numerous dynamic occlusions, such as moving pedestrians and vehicles. Many works have tried to address challenges in scenes featuring both static and transient objects.
RobustNeRF~\cite{Sabour2023RobustNeRFID} was the pioneer work that introduces a robust loss function that reduces the influence of photometrical inconsistency during training, thereby improving the overall reconstruction quality on its proposed dataset. Other methods~\cite{Chen2024NeRFHuGSIN, MartinBrualla2020NeRFIT, wu2022dnerf} also applied the similar idea to strengthen the robustness to outliers by down-weighting or discarding inconsistent observations based on the magnitude of color residuals.
NeRF On-the-go was the first to introduce DINOv2~\cite{oquab2024dinov} features to predict pixel uncertainty, in an attempt to achieve clean static reconstructions without transients.  It also introduced a benchmark dataset of casually captured images, aimed at evaluating static reconstruction performance for novel views. 

% \TODO{Remove WildGaussians}
Following these two methods and their datasets, recent methods of 3DGS have also explored techniques for separating transients and statics to reconstruct the statics more effectively. For example,
% WildGaussians~\cite{kulhanek2024wildgaussians} leveraged DINOv2~\cite{oquab2024dinov} features to model the uncertainty of bewteen the training and predicted images. 
SpotLessSplats~\cite{sabourgoli2024spotlesssplats} further extended previous methods by utilizing semantic features~\cite{oquab2024dinov} combined with robust optimization strategies to effectively eliminate transient distractors.
Besides, several approaches~\cite{MartinBrualla2020NeRFIT,kulhanek2024wildgaussians,Chen2021HallucinatedNR, FridovichKeil2023KPlanesER, kassab2023refinedfields, Yang2023CrossRayNR, Wang2024IENeRFIE, Chen2024NeRFHuGSIN, dahmani2024swag, Xu2024SplatfactoWAN, wang2024wegsinthewildefficient3d, zhang2024gaussian} have focused on reconstructing statics and illumination in datasets with unconstrained photo collections, that typically employ appearance embedding vectors to model the unique lighting and styles of each image, allowing for more accurate static reconstruction despite varying illumination status. 

Our method specifically focuses on distinguishing between transients and statics in casually captured unordered images with minimal illumination changes. Different from previous methods~\cite{Sabour2023RobustNeRFID, Ren2024NeRF, sabourgoli2024spotlesssplats}, we propose a hybrid Gaussian representation with 2D and 3D Gaussians with multi-view information to model the whole 3D scene without any additional semantic features.

\subsection{Image Representation with 2D Gaussians}
% Implicit Neural Representation (INR) is a widely used strategy for encoding 2D images by representing discrete data through continuous functions. It utilize neural networks to map pixel coordinates to RGB values, effectively embedding shapes, colors, and textures within the network parameters. However, the computational demands of these models often limit their usability on devices with constrained memory. 

Recently, inspired by 3DGS, some methods~\cite{zhang2024gaussianimage, zhang2024imagegscontentadaptiveimagerepresentation} in the field of image representation and compression have started to use 2D Gaussians to represent images. GaussianImage~\cite{zhang2024gaussianimage}
% , a novel alternative to CNNs, 
leverages 2D Gaussians on the pixel plane to achieve comparable reconstruction quality while offering enhanced compression and accelerated rendering capabilities. Image-GS~\cite{zhang2024imagegscontentadaptiveimagerepresentation} creates a content-adaptive image representation by fitting a target image by adaptively allocating and progressively optimizing a set of 2D Gaussians.

\textbf{Note that} the \textit{2D Gaussians} used here differ from the \textit{2D Surfels} (also referred to as \textit{2D Gaussians}) introduced by~\cite{Huang2DGS2024}. While 2D Surfels are designed to enable perspective-correct splatting through 2D surface modeling, ray-splat intersection, and volumetric integration in 3D space, and the 2D Gaussians in GaussianImage~\cite{zhang2024gaussianimage} is a flexible, compact, and content-adaptive image representation within 2D space.

Here, we incorporate the trainable 2D Gaussians to learn transient and uncertainty masks during the training process, decomposing the whole scene with 2D and 3D Gaussians.
\section{Preliminaries}
\label{sec:2d3dgs}

\textbf{3D Gaussian Splatting (3DGS)} represent a 3D scene explicitly with a set of 3D anisotropic Gaussians $ \{ \mathcal{G}_{3d} \} $. Each Gaussian is parameterized by its centroid $\mathbf{x_{3d}} \in \mathbb{R}^3$, scale $\mathbf{S_{3d}} \in \mathbb{R}^3$, rotation matrix  $\mathbf{R} \in {\mathbb{R}^{3 \times 3}}$, opacity $\mathbf{\alpha_{3d}} \in \mathbb{R}$ and color $\mathbf{c_{3d}} \in \mathbb{R}^3$ encoded in spherical harmonic ($\mathbf{SH}$) coefficients. The 3D covariance matrix $\mathbf{\Sigma_{3d}}$ of the 3D Gaussian is obtained from $\mathbf{R_{3d}}$ and $\mathbf{S_{3d}}$:
\begin{equation}
  \mathbf{\Sigma_{3d}} = \mathbf{R_{3d}}\mathbf{S_{3d}}\mathbf{S_{3d}}^T\mathbf{R_{3d}}^T
\end{equation} 
The 3D Gaussians are defined in world space for a spatial point $\mathbf{y_{3d}}$ following:
\begin{equation}
  \mathcal{G}_{3d}(\mathbf{y_{3d}}) = e^{-\frac{1}{2}(\mathbf{y_{3d}-x_{3d}})^T\mathbf{\Sigma_{3d}^{-1}}(\mathbf{y_{3d}-x_{3d}})}
\end{equation} 
Given $\mathbf{J}$, the Jacobian of the affine projective transformation and $\mathbf{W}$, a viewing transformation, the covariance matrix $\mathbf{\Sigma_{3d}'}$ in camera coordinates is defined as follows:
\begin{equation}
  \mathbf{\Sigma_{3d}'}= \mathbf{J}\mathbf{W}\mathbf{\Sigma_{3d}} \mathbf{W}^T\mathbf{J}^T
\end{equation}
These 3D Gaussians are projected to 2D imaging plane and blended through a fast, differentiable $\alpha$-blending process to render 2D images. The color $\mathbf{C_{3d}}$ of each pixel is computed using N-ordered 2D splats with the following formula:
\begin{equation}
 \mathbf{C_{3d}}=\sum_{i \in \mathbf{N}} \mathbf{\alpha_{3d}}_i' \mathbf{c_{3d}}_{i} \prod_{j=1}^{i-1}(1-\mathbf{\alpha_{3d}}_j')
\end{equation} 
where $\mathbf{\alpha_{3d}}_i'$ is the final opacity calculated by $\mathbf{\alpha_{3d}}_i'= \mathbf{\alpha_{3d}}_i * e^{-\frac{1}{2}(\mathbf{y_{3d}'}-\mathbf{x_{3d}'})^T\mathbf{\Sigma_{3d}'^{-1}}(\mathbf{y_{3d}'}-\mathbf{x_{3d}'})}$. And $\mathbf{x_{3d}'}$ and $\mathbf{y_{3d}'}$ are coordinates in the projected space. 
Typically, the centroids of 3D Gaussians are initialized using the sparse SfM point clouds obtained from the set of input images.

\begin{figure*}[t]
    \centering
    \includegraphics[width=1.0\linewidth]{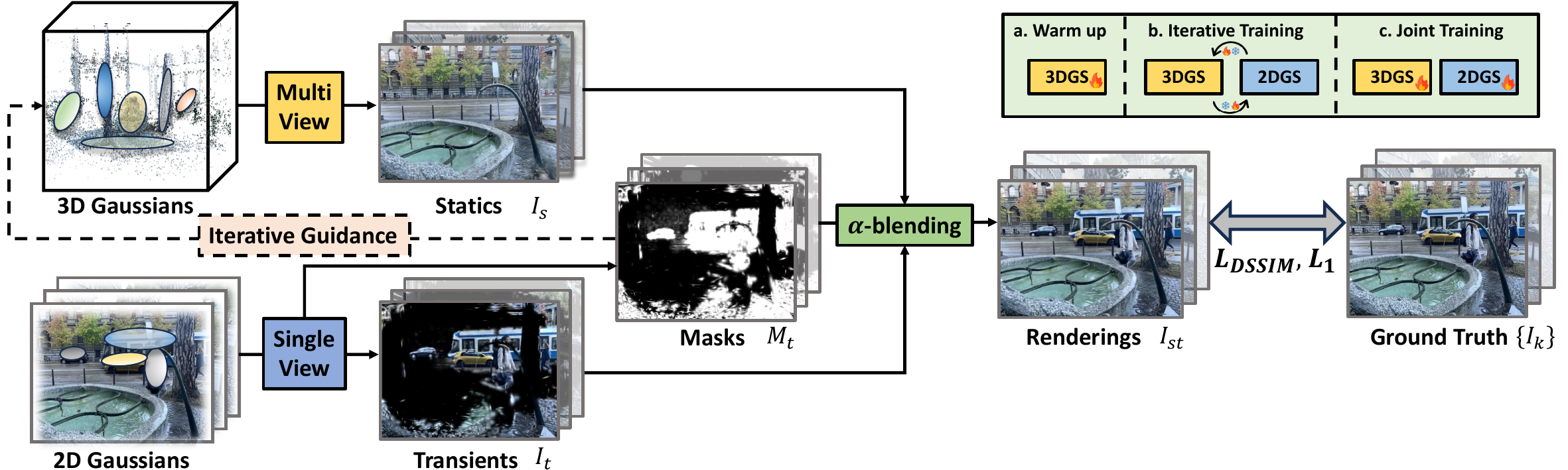}
    \caption{\textbf{Overview.} Given a casually captured image sequence, we decompose the whole scene into 2D Gaussians for transient objects and 3D Gaussians for static scenes. To warm up, we start by training a basic 3DGS to capture static elements. This is followed by iterative training of 2D and 3D Gaussians, where our transients and statics are combined using an $\alpha$-blending strategy with masks to produce the final renderings. The masks provide guidance for 3D Gaussians in the iterative training stage. During the joint-training, both 2D and 3D Gaussians are trained to further optimize the decomposition results.}
    \label{fig:overview}
\end{figure*}

\textbf{2D Gaussian Splatting (2DGS)} is initially introduced by ~\cite{zhang2024gaussianimage} in the task of implicit image representation. Since 2D Gaussians $ \{ \mathcal{G}_{2d} \} $ are no longer oriented to 3D scenes but to 2D space, many bloated operations and redundant parameters are discarded, such as viewing transformation, spherical harmonics, etc. Concretely, a basic 2D Gaussian is described by its centroid $\mathbf{x_{2d}} \in \mathbb{R}^2$, 2D covariance matrix $\mathbf{\Sigma_{2d}} \in \mathbb{R}^2 $, color $\mathbf{c_{2d}} \in \mathbb{R}^3$ and opacity $\mathbf{\alpha_{2d}} \in \mathbb{R}$. 
The 2D covariance matrix is also factorized into a rotation matrix $\mathbf{R_{2d}} \in {\mathbb{R}^{2 \times 2}}$ and a scale matrix $\mathbf{S_{2d}} \in \mathbb{R}^2$ as:
\begin{equation}
  \mathbf{\Sigma_{2d}} = \mathbf{R_{2d}}\mathbf{S_{2d}}\mathbf{S_{2d}}^T\mathbf{R_{2d}}^T
\end{equation} 
For rasterization, since the depth and camera parameters are not included in 2D Gaussians, a simplified accumulated summation of N-ordered 2D splats is calculated as follows: 
\begin{equation}
 \mathbf{C_{2d}} = \sum_{i \in \mathbf{N}} \mathbf{\alpha_{2d}}_i' \mathbf{c_{2d}}_i 
\end{equation} 
where $\mathbf{\alpha_{2d}}_i'$ is the opacity calculated by $\mathbf{\alpha_{2d}}_i'= \mathbf{\alpha_{2d}}_i * e^{-\frac{1}{2}(\mathbf{y_{2d}}-\mathbf{x_{2d}})^T\mathbf{\Sigma_{2d}^{-1}}(\mathbf{y_{2d}}-\mathbf{x_{2d}})}$. And $\mathbf{x_{2d}}$ and $\mathbf{y_{2d}}$ are coordinates of centroid and any pixel in the 2D image space.

Specifically, 2D Gaussians can be considered as projections of 3D Gaussians from a certain perspective, illustrating the relationship of Gaussians between 2D and 3D space. In our case, we combine the single-view independent 2D Gaussians and multi-view consistent 3D Gaussians to represent the entire 3D scene, which helps fundamentally decoupling transients and statics, based on the principles of multi-view geometry.

\section{Method}
\label{sec:method}

We delineate the pipeline of our approach as illustrated in Fig.~\ref{fig:overview}. Essentially, given a set of input images $\{I_k | k=1,2,..., N\}$ with the corresponding camera parameters, for each view $I$, the goal of our method is to reasonably decouple the transients $I_t$ and statics $I_s$ as follows:
\begin{equation}
    \label{eq:decompose}
    I = M_t \odot I_t + (1 - M_t) \odot I_s,
\end{equation}
where the $M_t \in [0, 1]$ represents the transient mask, and $\odot$ is per pixel multiplication. When a pixel's value in $M_t$ approaches 1, it indicates a higher probability that the location is transient; conversely, a lower value suggests a greater likelihood that area is static.

To achieve the goal, we decompose the whole 3D scene with two components: 
(1) \textbf{Multi-view consistent 3D Gaussians} is used for rendering $I_s$, which leverages multi-view information from images and models the static scene with a set of unified 3D Gaussians. The multi-view inputs regulate the 3D Gaussians to be consistent and robust across the different views.
(2) \textbf{Single-view independent 2D Gaussians} is responsible for modeling $I_t$, which enables our approach to handle the fact that images are casually captured with varying transients. Concretely, we forms a set of view-independent 2D Gaussians to model transients as planar objects from a single view.
This novel combination allows for a more precise and reasonable representation of 3D scenes, enabling better performance on novel views.

This section is organized as follows: we introduce our hybrid representation of scenes using 3D Gaussians in Sec.~\ref{subsec:3dgs} and 2D Gaussians in Sec.~\ref{subsec:2dgs}, followed by a comprehensive overview of the training pipeline in Sec.~\ref{subsec:training}.

\subsection{Modeling Statics with 3D Gaussians}
\label{subsec:3dgs}
Generally speaking, the initialized point clouds obtained by COLMAP~\cite{schoenberger2016sfm,schoenberger2016mvs} can only represent areas in the scene that satisfy multi-view consistency, so 3D Gaussians tend to reconstruct static scenes, such as buildings, grounds, etc. 

However, in scenes with transients, continuously training the 3D Gaussians may cause the transient objects to be overfitted into the Gaussian kernels because of constraints imposed by the RGB loss.
This approach fails to capture the true geometry and instead overemphasizes the RGB information, leading to annoying artifacts from other viewpoints and ultimately impacting quality of novel view synthesis.

We address this problem by adopting an enhanced multi-view regulated supervision scheme. 
Unlike previous methods that rasterize and supervise only a single image per iteration, our approach introduces two major changes.
(1) We increase the number of images per iteration to $K$, enabling gradient back-propagation to simultaneously consider mutual information from multiple views. This batch-wise inputs allow each optimization step to leverage multi-view insights to distinguish and infer transients and static elements. 
(2) We adopt a sparse training schedule that focuses optimization only on the 3D Gaussians within the co-visible areas of the cross-view frustums. This method not only sharpens the training focus but also reduces computational costs, as detailed in Alg.~\ref{alg:mv3dgs}, and outputs rasterized $\hat{I}_{s}$.

\begin{algorithm}[t]
    \caption{Multi-view Regulated Supervision.}
    \label{alg:mv3dgs}
    \SetAlgoLined
    \KwIn{A set of images $\{I_k, k=1,2,..., N\}$; Multi-view size $K$; Maximum iterations $T$;}
    \KwOut{Optimized 3D Gaussian parameters $\theta_{3d}$ within the co-visible regions;}
    Initialize $\theta_{3d}$ \;
    \For{each iteration $t = 1$ to $T$}{
    $I_{1,\ldots,K} \gets$ RandomSample($K$, $N$)\;
    $F_{\text{inter}} \gets \text{ComputeFrustum}(I_1)$\;
    \For{$i \gets 2$ \KwTo $K$}{
        $F_{\text{inter}} \gets F_{\text{inter}} \cap \text{ComputeFrustum}(I_i)$\;
    }
    $\mathcal{G}_{3d} \gets \text{ExtractGaussians}(I, F_{\text{inter}})$\;
    \For{each Gaussian $G_j \in \mathcal{G}_{3d}$}{
        $\theta \gets \theta - \eta \cdot \nabla \mathcal{L} (G_j(\theta))$\;
    }
    }
\end{algorithm}

\subsection{Modeling Transients with 2D Gaussians}
\label{subsec:2dgs}
Different from other methods that utilize semantic features, we introduce 2D Gaussians to model the transients as planar objects from a single view for each image. This approach is motivated by several key considerations: (1) 2DGS has been proven to be an efficient method for representing images, capable of modeling an arbitrary image effectively. (2) The formation and rasterization processes of 2DGS closely resemble those of 3DGS, allowing both to be expressed within a unified framework. (3) Degenerating 3D objects that do not satisfy multi-view geometric consistency into 2D representation can effectively decouple transients from static scenes on a fundamental level.

Specifically, 2D Gaussians are rasterized into an image $\hat{I}_{t}$ with a transient mask $\hat{M}_{t}$ for modeling uncertainty, each pixel $\mathbf{y_{2d}}$ of which is calculated as:
\begin{equation}
    \resizebox{0.9\hsize}{!}{$
    \hat{M}_{t}(\mathbf{y_{2d}}) = \sum_{i \in \mathbf{N}} \mathbf{\alpha_{2d}}_i' = \sum_{i \in \mathbf{N}} \mathbf{\alpha_{2d}}_i \cdot e^{-\frac{1}{2}(\mathbf{y_{2d}}-\mathbf{x_{2d}})^T\mathbf{\Sigma_{2d}^{-1}}(\mathbf{y_{2d}}-
    \mathbf{x_{2d}}).}
    $}
\end{equation}

It is noteworthy that we adopt uncertainty of 2D Gaussians to model the mask, rather than using 3D results. Because after modeling a static scene with 3D Gaussians, the 2D Gaussians are responsible for learning the residual part of the image. Therefore, naturally, the results rendered by the 2D Gaussian implicitly include a mask for transients.

\begin{table*}[ht]
    \begin{center}
    \caption{\textbf{Comparison on NeRF On-the-go Dataset.} For better visualization, the best results are highlighted in \hb{bold}. Our method shows generally superior performance over state-of-the-art methods. * indicates that the results are reproduced from the official implementation.}
    \label{table:onthego}
    \scalebox{0.6}{
    \begin{tabular}{c|ccc|ccc|ccc|ccc|ccc|ccc}
    \toprule
    \multirow{3}{*}{Method} & \multicolumn{6}{c|}{Low Occlusion}  & \multicolumn{6}{c|}{Medium Occlusion} & \multicolumn{6}{c}{High Occlusion} \\
    \multicolumn{1}{c|}{}  & \multicolumn{3}{c|}{Mountain}  & \multicolumn{3}{c|}{Fountain} & \multicolumn{3}{c|}{Corner}  & \multicolumn{3}{c|}{Patio} & \multicolumn{3}{c|}{Spot} & \multicolumn{3}{c}{Patio-High} \\
    % \midrule
    \cmidrule(lr){2-4} \cmidrule(lr){5-7} \cmidrule(lr){8-10} \cmidrule(lr){11-13} \cmidrule(lr){14-16} \cmidrule(lr){17-19}   
     & \psnr & \ssim & \lpips & \psnr & \ssim & \lpips & \psnr & \ssim & \lpips & \psnr & \ssim & \lpips & \psnr & \ssim & \lpips & \psnr & \ssim & \lpips \\
    \midrule
    RobustNeRF~\cite{Sabour2023RobustNeRFID} & 17.54 & 0.496 & 0.383 & 15.65 & 0.318 & 0.576 & 23.04 & 0.764 & 0.244 & 20.39 & 0.718 & 0.251 & 20.65 & 0.625 & 0.391 & 20.54 & 0.578 & 0.366 \\
    NeRF On-the-go~\cite{Ren2024NeRF} & 20.15 & 0.644 & 0.259 & 20.11 & 0.609 & 0.314 & 24.22 & 0.806 & 0.190 & 20.78 & 0.754 & 0.219 & 23.33 & 0.787 & 0.189 & 21.41 & 0.718 & 0.235 \\
    3DGS~\cite{Kerbl20233DGS}  & 19.40 & 0.638 & \hb{0.213} & 19.96 & 0.659 & \hb{0.185} & 20.90 & 0.713 & 0.241 & 17.48 & 0.704 & 0.199 & 20.77 & 0.693 & 0.316 & 17.29 & 0.604 & 0.363 \\
    SLS-mlp*~\cite{sabourgoli2024spotlesssplats} & 19.84 & 0.580 & 0.294 & 20.19 & 0.612 & 0.258 & 24.03 & 0.795 & 0.258 & 21.55 & \hb{0.838} & \hb{0.065} & 23.52 & 0.756 & \hb{0.185} & 20.31 & 0.664 & 0.259\\
    HybridGS(Ours) & \hb{21.73} & \hb{0.693} & 0.284 & \hb{21.11} & \hb{0.674} & 0.252 & \hb{25.03} & \hb{0.847} & \hb{0.151} & \hb{21.98} & 0.812 & 0.169 & \hb{24.33} & \hb{0.794} & 0.196 & \hb{21.77} & \hb{0.741} & \hb{0.211} \\
    \bottomrule
    \end{tabular}}
    \end{center}
    \vspace{-8pt}
\end{table*}

\begin{table*}[ht]
    \begin{center}
    \caption{\textbf{Comparison on RobustNeRF Dataset.} For better visualization, the best results are highlighted in \hb{bold}. Our method significantly outperforms all baseline methods. Crab (1) scene had a test set with same set of views as those in the train set, which is fixed in Crab (2).}
    \label{table:robustnerf}
    \scalebox{0.7}{
    \begin{tabular}{c|ccc|ccc|ccc|ccc|ccc}
    \toprule
    \multirow{2}{*}{Method} & \multicolumn{3}{c|}{Statue} & \multicolumn{3}{c|}{Android} & \multicolumn{3}{c|}{Yoda} & \multicolumn{3}{c|}{Crab (1)} & \multicolumn{3}{c}{Crab (2)}\\
    \cmidrule(lr){2-4} \cmidrule(lr){5-7} \cmidrule(lr){8-10} \cmidrule(lr){11-13} \cmidrule(lr){14-16}
    & \psnr & \ssim & \lpips & \psnr & \ssim & \lpips & \psnr & \ssim & \lpips & \psnr & \ssim & \lpips & \psnr & \ssim & \lpips \\
    \midrule
    RobustNeRF~\cite{Sabour2023RobustNeRFID} & 20.60 & 0.76 & 0.15  & 23.28  & 0.75 & 0.13 & 29.78 & 0.82  & 0.15 & 32.22 & 0.94 & 0.06 & - & - & - \\
    NeRF On-the-go~\cite{Ren2024NeRF} & 21.58  & 0.77 & 0.24  & 23.50  & 0.75  & 0.21 & 29.96 & 0.83 & 0.24 & - & - & - & - & - & - \\
    3DGS~\cite{Kerbl20233DGS}  & 21.02  & 0.81 & 0.16 & 23.11 & 0.81 & 0.13 & 26.33 & 0.91  & 0.14 & 31.80 & 0.96  & 0.08  & 29.74 & - & - \\
    SLS-mlp~\cite{sabourgoli2024spotlesssplats} & 22.54 & 0.84 & 0.13 & 25.05 & \hb{0.85} & 0.09 & 33.66 & \hb{0.96}  & 0.10 & 35.85 & \hb{0.97}  & 0.08  &  34.43 & - & - \\
    HybridGS(Ours)  & \hb{22.93} & \hb{0.87} & \hb{0.10} & \hb{25.15} & \hb{0.85} & \hb{0.07} & \hb{35.32} & \hb{0.96} & \hb{0.07} & \hb{36.31} & \hb{0.97} & \hb{0.05} & \hb{35.17} & \hb{0.96} & \hb{0.08} \\
    \bottomrule
    \end{tabular}}
    \end{center}
    \vspace{-8pt}
\end{table*}

\subsection{Multi-stage Training Scheme}
\label{subsec:training}
Because 2D and 3D Gaussians are trained within the same framework, it is crucial to balance their relationship effectively. To that end, we have proposed a multi-stage training strategy that ensures stable convergence and enhances overall performance.

\textbf{Warm up Pre-training.} 
The training process begins by using 3DGS to capture the essential structure of the entire static scene. This stage focuses on modeling the scene with a set of unified 3D Gaussians with our multi-view regulated supervision as detailed in Sec.~\ref{subsec:3dgs}, producing foundational, albeit low-quality, rendered images ${\hat{I}_{s}}$ using a combination of DSSIM~\cite{Wang2004ImageQA} and L1 losses following:
\begin{equation}
\begin{aligned}
    \mathcal{L}_{warmup} &= \lambda L_{DSSIM}(\hat{I}_{s}, I) + (1-\lambda) L_1(\hat{I}_{s}, I).
\end{aligned}
\label{eq:loss_warmup3d}
\end{equation}
where the $\lambda$ is set to 0.2. During the warm-up, the densification of 3D Gaussians is conducted to gradually increase the number, allowing for a more detailed representation of the static scene structure. 

\textbf{Iterative Training.} 
The second stage employs iterative training between 2DGS and 3DGS to progressively optimize both components. 
During the alternation, the 3DGS will provide more accurate static scene renderings, which in turn enables the 2DGS to refine its transient representations ${\hat{I}_{t}}$ and mask predictions ${\hat{M}_{t}}$ with Eq.~\ref{eq:loss_warmup2d}. 
Conversely, the improved masks from 2DGS allow the 3DGS to focus more precisely on rendering the static regions of the scene with Eq.~\ref{eq:loss_3dgs}. The rendered image $\hat{I}$ is a composition of $\hat{I}_s$ and $\hat{I}_t$ with $\hat{M}_t$ following Eq.~\ref{eq:decompose}. This process iteratively refines the transients and statics, improving the overall quality and accuracy of the masks. The loss functions are as follows:
\begin{equation}
\begin{aligned}
    \mathcal{L}_{iter2d} &= \lambda L_{DSSIM}(\hat{I}, I) + (1-\lambda) L_1(\hat{I}, I),
\end{aligned}
\label{eq:loss_warmup2d}
\end{equation}
\begin{equation}
\begin{aligned}
    \mathcal{L}_{iter3d} &= (1-\hat{M}_{t}) \odot \mathcal{L}_{iter2d} .
\end{aligned}
\label{eq:loss_3dgs}
\end{equation}
Note that, during training, 2DGS and 3DGS are trained alternately. When training one branch, the gradient backpropagation of the other branch is turned off.

\textbf{Joint Fine-tuning.} 
The final stage involves joint fine-tuning, integrating 2DGS and 3DGS into a unified framework. 
This stage focuses on fine-tuning on the transient mask and reducing the error between the rendered image $\hat{I}$, and the ground truth image $I$. 
The loss $\mathcal{L}_{joint}$ used during the joint fine-tuning is a combination as follows:
\begin{equation}
\begin{aligned}
    \mathcal{L}_{joint} &= \beta L_{DSSIM}(\hat{I}, I) + (1-\beta) L_1(\hat{I}, I),
\end{aligned}
\label{eq:loss_st}
\end{equation}
where the $\beta$ is set to 0.2. Here, both the 2DGS and 3DGS branches will enable gradient backpropagation to simultaneously update all parameters.

\section{Experiments}
\label{sec:exp}

\subsection{Experimental Setup}

\textbf{Datasets.} Similar to previous methods, we evaluate our HybridGS on two challenging datasets: NeRF On-the-go~\cite{Ren2024NeRF} and RobustNeRF~\cite{Sabour2023RobustNeRFID}. NeRF On-the-go contains multiple casually captured indoor and outdoor sequences with varying ratios of distractors (from 5\% to over 30\%). We use the version of the dataset where all images were undistorted as ~\cite{kulhanek2024wildgaussians}. The RobustNeRF dataset includes several scenes exemplifying different types of distractors. To simulate capturing over extended periods, distractor objects are repositioned between frames to interfere with statically placed objects (from 1 (Statue) to 150 (Yoda)). The splits includes distractor-filled training set and distractor-free testing set.

\noindent \textbf{Implementation Details.} Our HybridGS, developed on top of open-source implementation gsplat~\cite{ye2024gsplatopensourcelibrarygaussian} and Taming-3DGS~\cite{taming3dgs}, incorporates custom CUDA kernels, boosting the rasterization of $\alpha$-blending for 3DGS and weighted summation for 2DGS. The number of views $K$ is 4 in our practice.
For a scene consisting of 100 images, the process starts with a 1k-step warm-up. This is followed by iterative training, which includes 10k steps for 2DGS and 1k steps for 3DGS.  The final joint training involves 30k steps. 
The entire training process takes approximately 0.18 GPU hours. The number of training steps scales proportionally with the size of the dataset. All experiments are performed on a single NVIDIA RTX 4090 GPU using PyTorch.

\noindent \textbf{Metrics.} We follow common practice by employing PSNR, SSIM~\cite{Wang2004ImageQA}, and LPIPS~\cite{Zhang2018TheUE} for evaluation. 

\subsection{Evaluation}

\begin{table}[!t]
    \begin{center}
    \caption{\textbf{Ablation on components in datasets with varying occlusions.} The best results are highlighted in \hb{bold}.}
    \label{table:components}
    \scalebox{0.55}{
    \begin{tabular}{c|ccc|ccc|ccc}
    \toprule
    \multirow{2}{*}{Dataset(ratio)}  & \multicolumn{3}{c|}{On-the-go low.(5\%)} & \multicolumn{3}{c}{On-the-go medium.(17\%)} & \multicolumn{3}{|c}{On-the-go high.(26\%)} \\
    \cmidrule(lr){2-4} \cmidrule(lr){5-7} \cmidrule(lr){8-10}
    & \psnr & \ssim & \lpips & \psnr & \ssim & \lpips & \psnr & \ssim & \lpips \\
    \midrule
    Ours & \hb{21.42} & 0.684 & \hb{0.268} & \hb{23.50} & \hb{0.827} & \hb{0.167} & \hb{23.05} & \hb{0.768} & 0.203 \\
    w/o iterative Training  & 21.39 & \hb{0.687} & 0.269  &  23.11 & 0.815 & 0.174 & 22.55 & 0.753 & 0.218 \\ 
    w/o Joint Training & 21.05 & 0.654 & 0.305 & 23.18 & 0.815 & 0.179 & 23.01 & 0.748 & 0.223 \\
    w/o Multi-view & 21.02 & 0.658 & 0.300 & 23.06 & 0.812 & 0.176 & 22.71 & 0.763 & \hb{0.202} \\ 
    \bottomrule
    \end{tabular}}
    \end{center}
\end{table}

\subsubsection{Comparison on the NeRF On-the-go Dataset}
\begin{figure*}[th]
    \includegraphics[width=\linewidth]{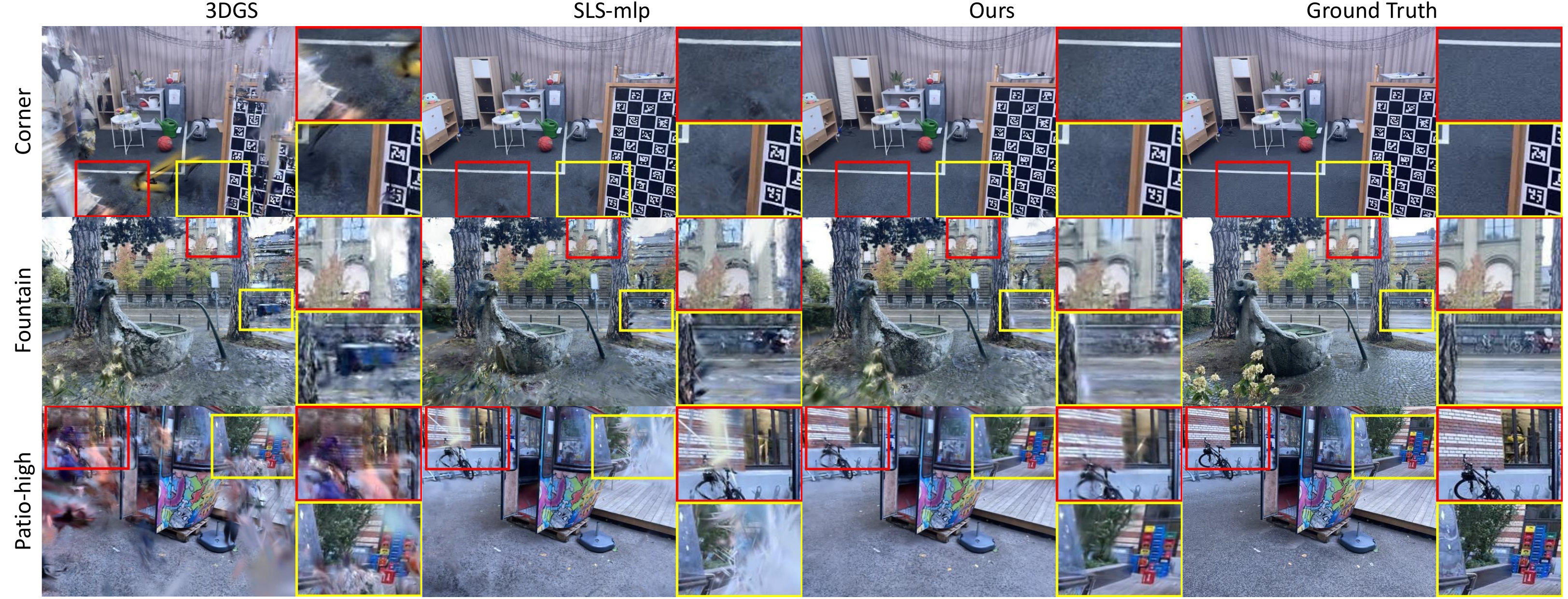}
    \caption{\textbf{Visualization of novel view synthesis results on the testing set of NeRF On-the-go dataset.} Our method demonstrates superior results by effectively reducing artifacts and providing clearer boundaries. This results in a cleaner statics compared to other methods, showcasing enhanced visual quality and precision in novel views.}
    \label{fig:nvs}
\end{figure*}

As shown in Tab.~\ref{table:onthego} and Fig.~\ref{fig:nvs}, our method demonstrates superior performance across different scenarios with varying levels of occlusion. 
In Tab.~\ref{table:onthego}, our method outperforms the previous SOTA method across multiple metrics, with an average improvement of 1.10 dB in PSNR and 5.27\% in SSIM, demonstrating the superiority of our approach.

Furthermore, Fig.~\ref{fig:nvs} provides a qualitative comparison, illustrating the visual fidelity of our results against other methods. Our method produces more clear and detailed images, closely resembling the ground truth, particularly in challenging scenes with complex geometry and occlusions. This highlights our model's capability to effectively handle diverse visual conditions both in indoor and outdoor scenes.

\subsubsection{Comparison on the RobustNeRF Dataset}
\begin{figure}[t]
    \centering
    \includegraphics[width=\linewidth]{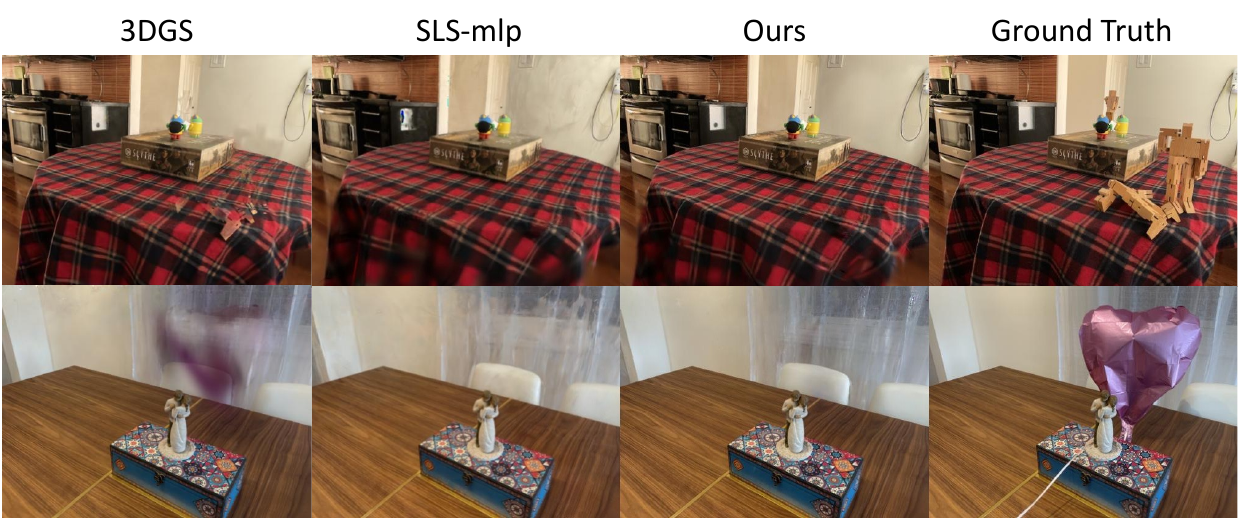}
    \caption{\textbf{Visualization of novel view synthesis results on RobustNeRF dataset.}}
    \label{fig:robustnerf}
\end{figure}

We present the quantitative results of RobustNeRF in Tab.~\ref{table:robustnerf}. Our method demonstrates superior performance quantitatively compared to all previous approaches. 
In scenes with numerous transients, such as the Yoda scene, we've noticed that the performance metrics of the original 3DGS significantly degrade, occasionally even dropping below those of NeRF-related methods.
By leveraging multi-view shared information and incorporating 2D Gaussians to effectively represent transients, our approach surpasses semantic-based methods like SLS-mlp~\cite{sabourgoli2024spotlesssplats}. As shown in Fig.~\ref{fig:robustnerf}, our results have clearer overall boundaries and structured fine-grained details, while SLS-mlp~\cite{sabourgoli2024spotlesssplats} may lead to blurriness in static objects and occluded areas. This further confirms that, even when faced with numerous transients, our method consistently maintains stable and reliable performance.

\subsection{Ablation Studies}

\begin{table}[!t]
    \begin{center}
    \caption{\textbf{Ablation on the number of 2D Gaussians of Corner in NeRF On-the-go dataset.} For better visualization, the best results are highlighted in \hb{bold}.}
    \label{table:2dgs}
    \scalebox{0.6}{
    \begin{tabular}{c|ccc|cc}
    \toprule
    \multirow{2}{*}{Number} & \multirow{2}{*}{\psnr} & \multirow{2}{*}{\ssim} & \multirow{2}{*}{\lpips} & \multicolumn{2}{c}{Storage (MB)} \\
    & & & & 2DGS & 3DGS \\
    \midrule
    1k & 24.493 & 0.840 & \hb{0.150} & 3.5 & 42.1 \\
    5k & 24.733 & 0.843 & 0.158 & 17.3 & 41.9 \\ 
    10k & \hb{25.034} & \hb{0.847} & 0.151 & 34.7 & 42.2 \\
    15k & 24.512 & 0.839 & 0.167 & 42.6 & 52 \\
    25k & 24.698 & 0.840 & 0.167 & 69.4 & 42.2 \\
    \bottomrule
    \end{tabular}}
    \end{center}
    \vspace{-18pt}
\end{table}

\noindent \textbf{Components.} 
To assess the effectiveness of each component in our framework, we performed an ablation study, as presented in Tab.~\ref{table:components}. We systematically removed iterative training, joint training, and multi-view supervision to evaluate their effectiveness. Each component, as expected, enhances overall performance, demonstrating their individual and combined significance. 
In scenes with low occlusion, multi-view supervision allows the model to concentrate on static elements by integrating data from various perspectives. 
Conversely, in highly occluded environments, joint training with 2D and 3D Gaussians markedly improves performance, highlighting the model’s capability to manage complex scenes with significant occlusions and transient changes. 
Iterative training consistently boosts results across different occlusion levels.
In summary, the results underscore that integrating diverse training strategies and multi-view information significantly enhances the rendering quality of unseen views, especially in challenging conditions.

\begin{figure}[t]
    \centering
    \includegraphics[width=\linewidth]{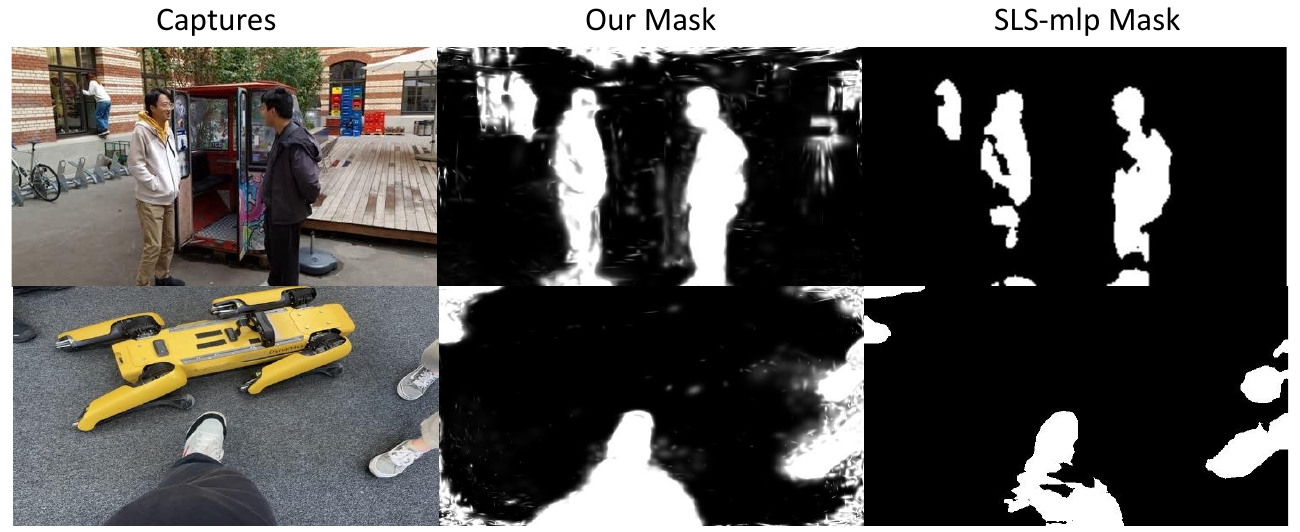}
    \caption{\textbf{Comparison of transient masks on NeRF On-the-go dataset.}}
    \label{fig:comparemask}
\end{figure}

\begin{figure*}[th]
    \centering
    \includegraphics[width=\linewidth]{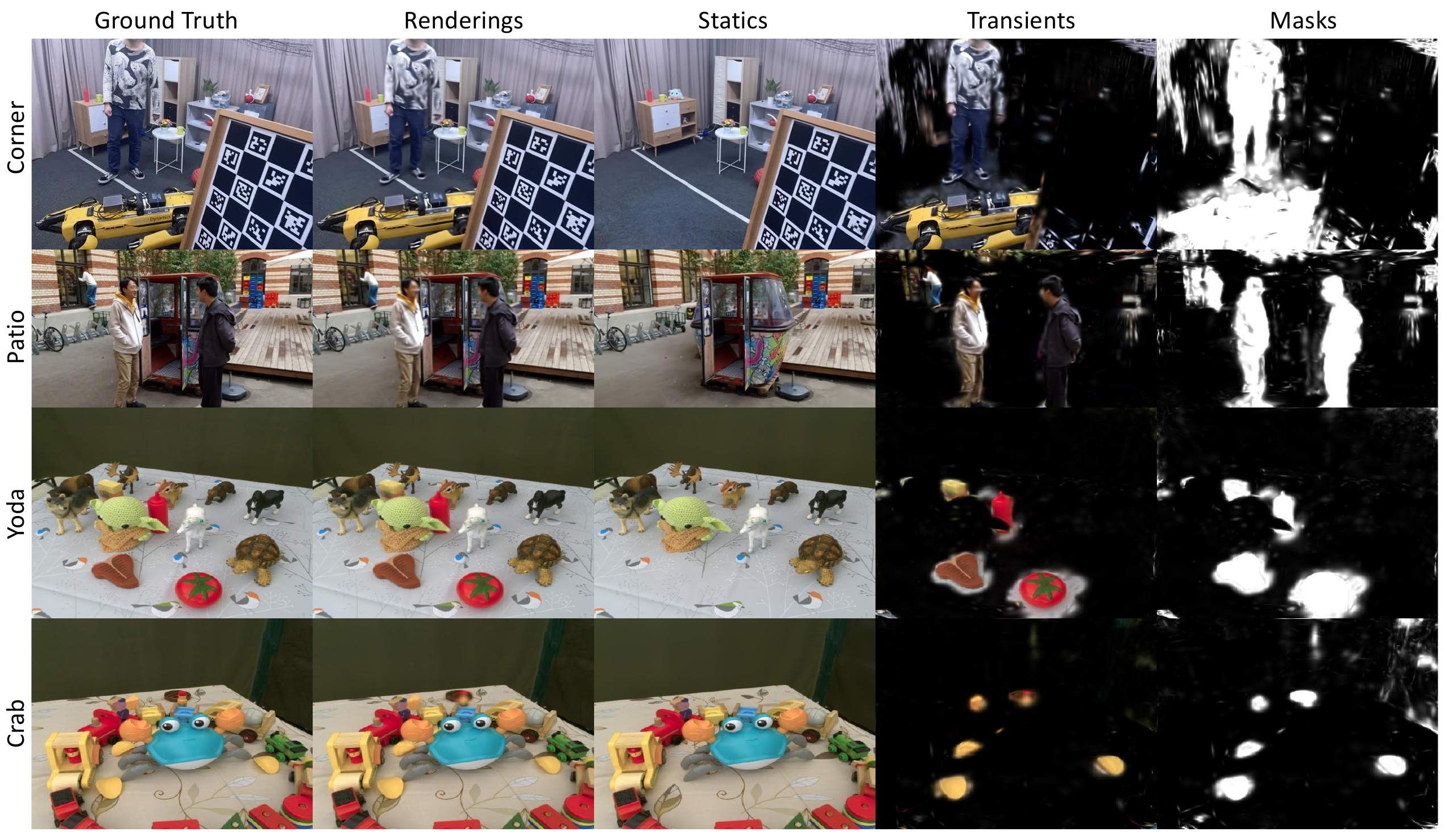}
    \caption{\textbf{Visualization of scene decomposition into transients and statics.} Our method achieves superior transient mask separation in both indoor and outdoor scenes. It effectively separates transients and statics, and the resulting renderings closely resemble the ground truth images, demonstrating its effectiveness. Top two rows are from NeRF On-the-go~\cite{Ren2024NeRF}, bottom two rows are from RobustNeRF~\cite{Sabour2023RobustNeRFID}.}
    \label{fig:mask}
    % \vspace{-8pt}
\end{figure*} 

\noindent \textbf{The Number of 2D Gaussians.} Tab.~\ref{table:2dgs} presents an ablation study on the number of 2D Gaussians used in the corner sequence of NeRF On-the-go dataset. The results indicate that increasing the number of 2D Gaussians initially improves the performance. Specifically, 10k 2D Guassians achieve the highest PSNR (25.034) and SSIM (0.847), while maintaining a competitive LPIPS score of 0.151 and reasonable storage requirements. Based on these findings, we select 10k as the optimal number in our experiments.

\noindent \textbf{Transients and Masks.}  Benefiting from our method's explicit modeling of transient objects, we can obtain transient masks without introducing any segmentation networks or other pre-trained features. Typically, these masks capture dynamic elements such as pedestrians and vehicles, as well as some structural details ignored by 3D Gaussians. In Fig.~\ref{fig:mask}, our approach effectively learns the RGB and mask of transients even under strong occlusions, while maintaining the robustness of static elements. This capability underscores the strength of our method in discerning and isolating dynamic components within complex scenes. 

We also compare our transient masks with SLS-mlp~\cite{sabourgoli2024spotlesssplats} semantic masks in Fig.~\ref{fig:comparemask}.
It is observed that the semantic mask does not perfectly align with the transients
% . This misalignment is partly 
due to the semantic features not being sufficiently precise. Additionally, transients often encompass low-level information, such as motion blur or object shadows, which may not explicitly exhibit clear semantic features. By modeling these transients using 2D Gaussians at the pixel level, our method allows for a more fine-grained capture of transients.

\begin{figure}[t]
    \includegraphics[width=\linewidth]{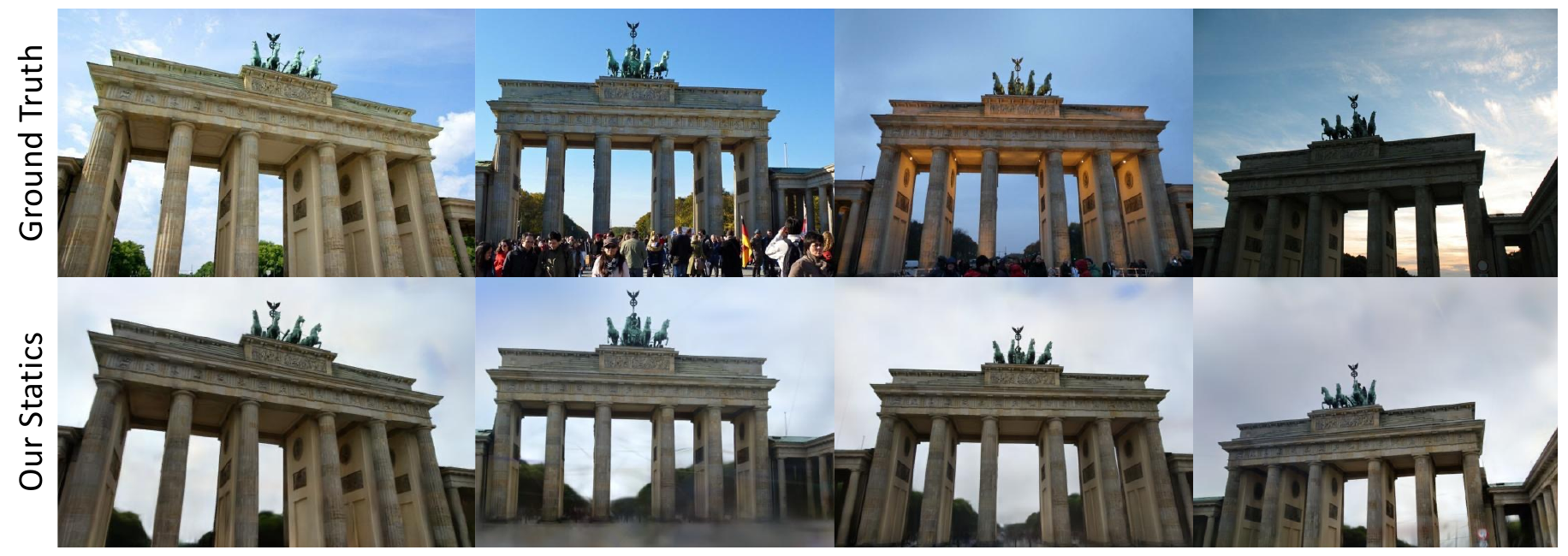}
    \caption{\textbf{Visualization on Photo Tourism dataset.}}
    \label{fig:photo}
    % \vspace{-20pt}
\end{figure}

\subsection{Discussions}
While our method has demonstrated strong performance in distinguishing between transients and statics, it has not yet accounted for the illumination variation in unconstrained photo collections. 
As illustrated in Fig.~\ref{fig:photo} on Photo Tourism~\cite{SSS:2006}, our method renders fine-grained architectural structures, but the overall photometric result tends towards an average photometric effect, akin to overcast conditions. A promising direction is to incorporate an appearance embedding module in our method for the future work.

\section{Conclusion}
\label{sec:conclusion}

In this paper, we introduce HybridGS, a novel hybrid representation with 2D and 3D Gaussians for decomposing 3D scenes from casually captured images. 
By introducing the hybrid Gaussian representation, our method addresses the challenges of handling transients as planar objects while preserving the integrity of the static scene. 
The multi-view supervision mechanism we developed plays a crucial role in enhancing the discrimination between static and transient elements across multiple views, reducing artifacts and inaccuracies in novel views. 
We have developed a multi-stage training strategy for the joint training of 2D and 3D Gaussians, ensuring the stable and robust training stage. 
Our experiments demonstrate that HybridGS significantly outperforms existing methods in terms of quality and efficiency.
In summary, HybridGS establishes a novel, robust and efficient foundation for 3D scene representation, paving the way for advancements in various vision-based applications.

{
    \small
    \bibliographystyle{ieeenat_fullname}
    \bibliography{main}
}

\clearpage
\setcounter{page}{1}
\maketitlesupplementary

\section{Positioning of Our Work}
\label{sec:background}
In the community, there are currently two predominant approaches for tackling the challenge of novel view synthesis in wild images with different complexities. We distinguish these approaches according to the primary datasets they utilize and provide a detailed comparison in Tab.~\ref{table:comparison}.

NeRF On-the-go~\cite{Ren2024NeRF} processes casually captured images that lack inter-frame continuity, with the goal of eliminating the interference from transient objects to reconstruct statics. 

Photo Tourism~\cite{SSS:2006} gathers photo collections from the web, resulting in completely unconstrained conditions with more complex lighting variations and increased foreground interference. It focuses more on integrating appearance embedding to model the photometric changes in the scene.

In summary, our method belongs to the first category and aims to decompose transients and statics from casually captured images in scenes with minimal illumination changes. Experiments have demonstrated our state-of-the-art results on two widely used benchmark datasets, such as NeRF On-the-go~\cite{Ren2024NeRF} and RobustNeRF~\cite{Sabour2023RobustNeRFID}. Handling varying lighting conditions will be our future work as discussed in the paper.

\section{More Discussions}
\subsection{2D Gaussians}
\label{subsec:2dgs}

The fitting capability of 2D Gaussians is inherited from 3D Gaussians. Given $\mathbf{J}$, the Jacobian of the affine projective transformation, and $\mathbf{W}$, the viewing transformation, the 3D Gaussians can be projected to 2D image plane and blended through a fast, differentiable $\alpha$-blending process to render 2D images following the Eq. 3 and Eq. 4. Therefore, the 2D Gaussians can be viewed as the projection of 3D Gaussians. 

During training, the warm-up allows 3DGS to establish an initial model of the entire scene. It is noteworthy that intuitively, the residuals between the results of 3DGS rendering and the ground-truth would potentially model transients. However, the 3DGS itself is constrained only by RGB loss, therefore, the transient objects from different viewpoints are eventually fitted into the 3D Gaussians, leading to less effective fitting of static scenes with vanilla 3DGS. We address this issue by incorporating additional 2D Gaussians. During the iterative training stage, the 2DGS learns the residuals per view, focusing more on the unique elements of each image. The output soft mask or matting can effectively direct 3DGS to concentrate on areas with smaller residuals, which represent the common and shared elements of the scene. In the final joint training stage, we perform a deep integration of 2D and 3D Gaussians for fine-tuning.
Therefore, in our method, 3D Gaussians tend to learn the elements that are consistent across different viewpoints, which we define as statics. Meanwhile, 2D Gaussians capture image-specific information, such as dynamic objects and occlusions, referred to as transients.

\begin{table}[t]
    \begin{center}
    \caption{\textbf{Comparison of NeRF On-the-go and Photo Tourism.}}
    \label{table:comparison}    
    \scalebox{0.71}{
    \begin{tabular}{c|cc}
    \toprule
    \textbf{~} & \textbf{NeRF On-the-go}~\cite{Ren2024NeRF} & \textbf{Photo Tourism}~\cite{SSS:2006} \\
    \hline
    Data source  & Casually captured photos & Web photos in the wild \\
    \hline
    Photometric & Similar lighting & Varying lighting over time \\
    \hline
    Scene & Indoor and outdoor scenes & Mostly outdoor scenes \\
    \hline
    Evaluation  & Statics & Statics with their illumination \\
    \hline
    Related Works  & ~\cite{Ren2024NeRF, kulhanek2024wildgaussians, sabourgoli2024spotlesssplats} \& \textbf{Our HybridGS} & ~\cite{MartinBrualla2020NeRFIT, Chen2021HallucinatedNR, Chen2024NeRFHuGSIN, dahmani2024swag, wang2024wegsinthewildefficient3d, zhang2024gaussian} \\
    \bottomrule
    \end{tabular}}
    \end{center}
\end{table}

\subsection{Multi-view Supervision}
\label{subsec:mv}

\begin{table}[!t]
    \begin{center}
    \caption{We conduct ablation studies on a pure static scene \textit{Garden} from the MipNeRF 360 dataset~\cite{barron2022mipnerf360}, and \textit{Corner},  scene that includes both transients and statics, from the NeRF On-the-go dataset~\cite{Ren2024NeRF}, to explore the potential influence of our designs.}
    \label{table:multiview}
    \scalebox{0.73}{
    \begin{tabular}{c|ccc|ccc}
    \toprule
    \multirow{2}{*}{Method Settings} & \multicolumn{3}{c|}{Garden} & \multicolumn{3}{c}{Corner} \\
    & \psnr & \ssim & \lpips & \psnr & \ssim & \lpips \\ 
    \midrule
    3DGS  & 29.323 & 0.924 & 0.050 & 20.148 & 0.686 & 0.290 \\
    + Multi-view & 29.572 &  0.925 &  0.053 &  21.758 & 0.769 &  0.235 \\
    + Multi-view + 2DGS & 29.512 & 0.924 & 0.054 &  25.020  & 0.842 &  0.165 \\
    \bottomrule
    \end{tabular}}
    \end{center}
\end{table}

\begin{figure*}
  \centering
  \includegraphics[width=1.0\textwidth]{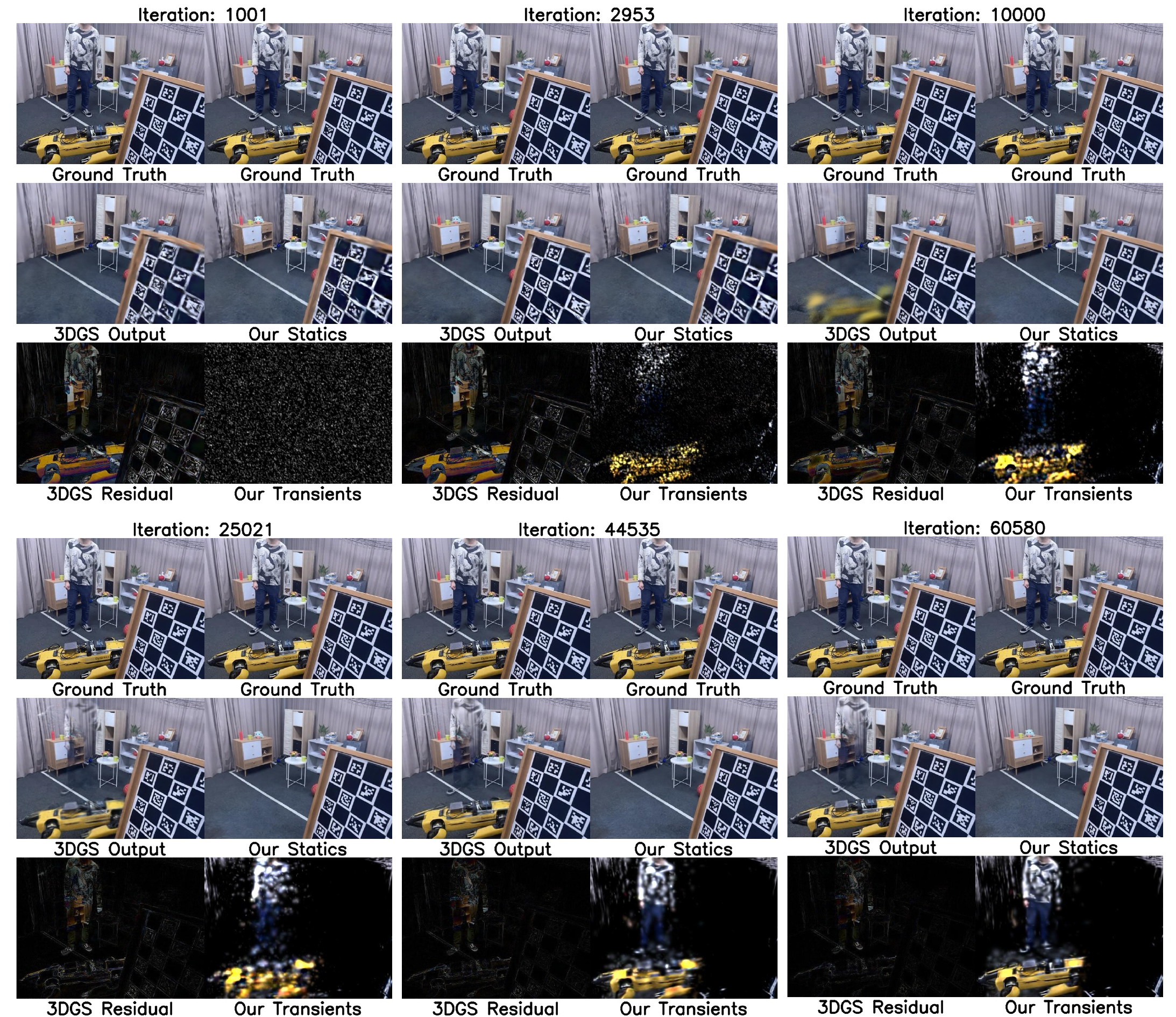}
  \caption{\textbf{Qualitative results compared to 3DGS at each randomly selected training step.} For HybridGS, the training steps for each stage are: Warm-up: 0$\sim$1,010, Iterative Training: 1,010$\sim$40,400, Joint Training: 40,400$\sim$60,600. Notably, this scene contains 101 images. Referring to Fig.~\ref{fig:psnr}, before the warm-up step (the top left), during the initial rough training phase, both our and 3DGS results are somewhat blurry. However, since we adopt a multi-view strategy, the occluded parts in our results are slightly clearer. As training progresses, by the 2953rd iteration (the top central), 3DGS reaches its optimum. Nevertheless, at this point, the background in the transients remains quite blurry for 3DGS, whereas our approach has already transitioned into the iterative training phase, allowing us to model static elements more accurately. 
  Moving forward, we maintain stable training (top right and bottom left), largely due to our introduction of 2D Gaussians to decouple transients from statics. This effectively prevents over-fitting to transients that 3DGS begins to experience, leading to diminishing rendering quality.
  By the 44535th iteration (the bottom central), during the joint optimization phase, our results reach their optimum. The bottom right shows the results at the end of the training process.
  }
  \label{fig:training}
\end{figure*}

\begin{figure*}
  \centering
  \includegraphics[width=1.0\textwidth]{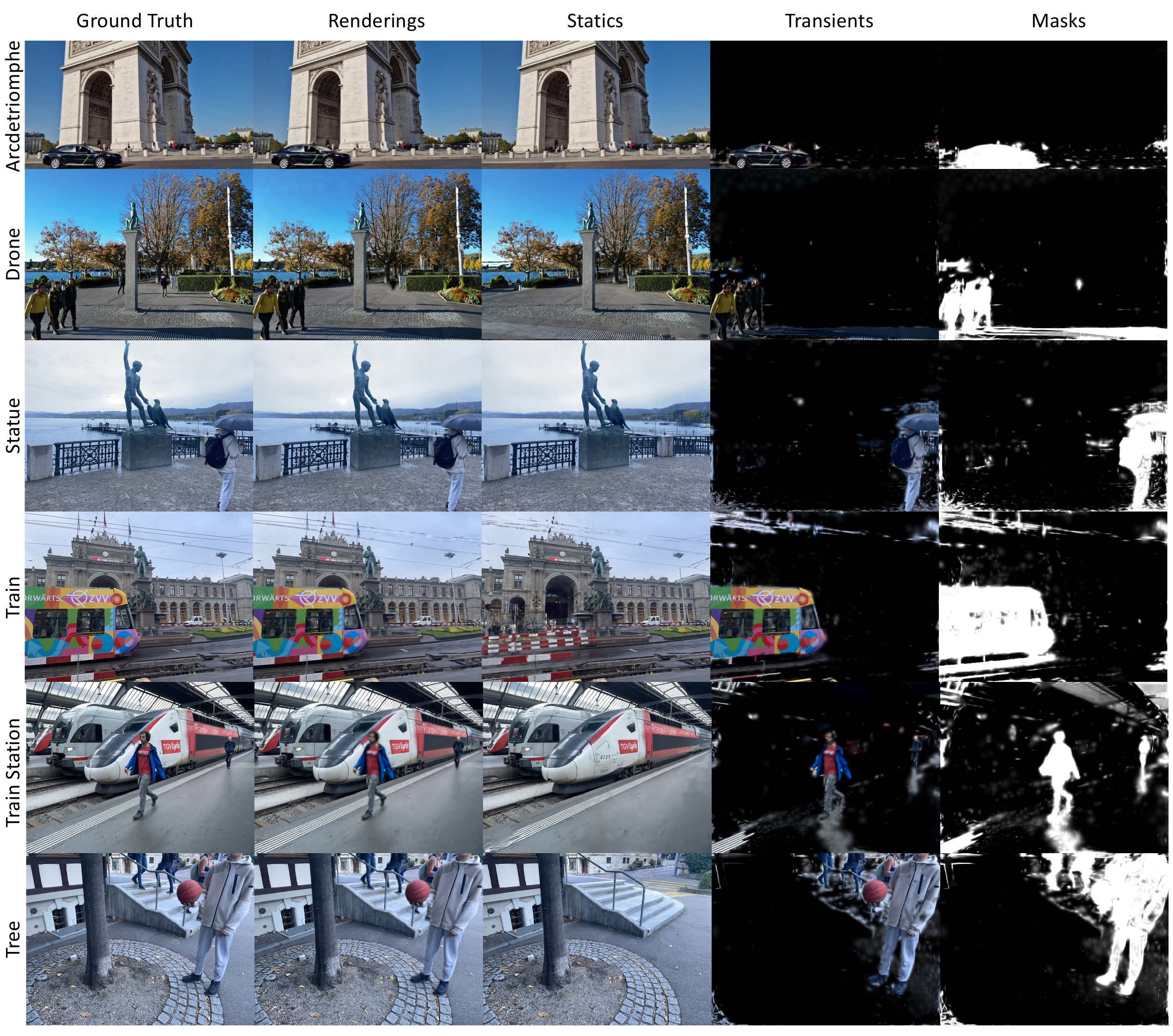}
  \caption{\textbf{Visualization on 6 remaining scenes of NeRF On-the-go~\cite{Ren2024NeRF} dataset.}}
  \label{fig:on-the-go}
\end{figure*}

We have further investigated the performance of multi-view 3DGS in different scenarios in Tab.~\ref{table:multiview}. To be specific, we select the static scene \textit{Garden} from the MipNeRF 360~\cite{barron2022mipnerf360} dataset and the dynamic scene \textit{Corner} from the NeRF On-the-go~\cite{Ren2024NeRF} dataset. The results indicate that in static scenes, employing multi-view supervision has minimal impact on achieving the best results for novel view synthesis. However, during training, it is noted that 3DGS tends to over-fit the training views, leading to a gradual decline in both visual quality and metric performance for novel views. In contrast, multi-view 3DGS demonstrates more stable convergence and effectively reduces the over-fitting problem. In dynamic scenes, obviously, 3DGS is prone to over-fitting, which adversely affects performance in novel view synthesis. On the other hand, multi-view 3DGS benefits from mutual supervision in areas visible to multiple views, significantly improving the visual results.

\section{More Implementation Details}
\label{sec:imple}
\subsection{Training}

For the training of 3D Gaussians, we perform the densification of 3D Gaussians during the warm-up stage. Then, in the subsequent stages, we maintain a constant number of existing 3D Gaussians and focus solely on optimizing their parameters.
For 2D Gaussians, we maintain a constant number 10,000 per image throughout the entire training process without any densification. 
During the iterative training process, while optimizing 3DGS with 2DGS held fixed, we binarize the uncertainty mask obtained from 2DGS into 0s and 1s using a threshold value of $\epsilon=0.1$.

\subsection{Datasets}

We follow the same training/testing split and resolution settings as the official rules in NeRF On-the-go~\cite{Ren2024NeRF} and RobustNeRF~\cite{Sabour2023RobustNeRFID}. 
Specifically, for the NeRF On-the-go dataset, we downsample images from most scenes by \(8\times\) to \(504 \times 378\). Note that \textit{Arcdetriomphe} and \textit{Patio} are downsampled by \(4\times\) to \(480 \times 270\). For the RobustNeRF dataset, all scenes are downsampled by \(8\times\), with \textit{Android} and \textit{Statue} resized to \(503 \times 377\), and \textit{Crab} and \textit{Yoda} to \(431 \times 431\).
\subsection{Storage}

\begin{table}[!t]
    \begin{center}
    \caption{\textbf{Comparison of storage and computational efficiency during the training and testing.} We provide a comprehensive statistical evaluation on the \textit{Corner} of NeRF On-the-go dataset.}
    \label{table:scenecompression}
    \scalebox{0.7}{
    \begin{tabular}{c|cc|cc|c}
    \toprule
    \multirow{2}{*}{Method} & \multicolumn{2}{c|}{Storage (MB)} &  \multicolumn{2}{c|}{Training} & Testing \\
    &  3DGS & 2DGS & Time (Hours) & Memory (GB) & FPS (Hz) \\
    \midrule
    3DGS~\cite{Kerbl20233DGS}  &  410.70 & 0 &  0.35 & 3.21  &  116 \\
    Ours   & 42.20 & 34.70 & 0.18 & 2.20 & 160 \\
    \bottomrule
    \end{tabular}}
    \end{center}
\end{table}

Tab.~\ref{table:scenecompression} highlights the advantages of our method in terms of storage and computational efficiency. Compared to 3DGS, our method use 2D Gaussians to model transients instead of forcing 3D Gaussians to fit them, significantly reducing both storage and computational requirements, showcasing its superior efficiency in both training and testing phases.

\section{More Visualization Results}
\subsection{Training Process}\label{subsec:training_process}
\begin{figure}
  \centering
  \includegraphics[width=0.48\textwidth]{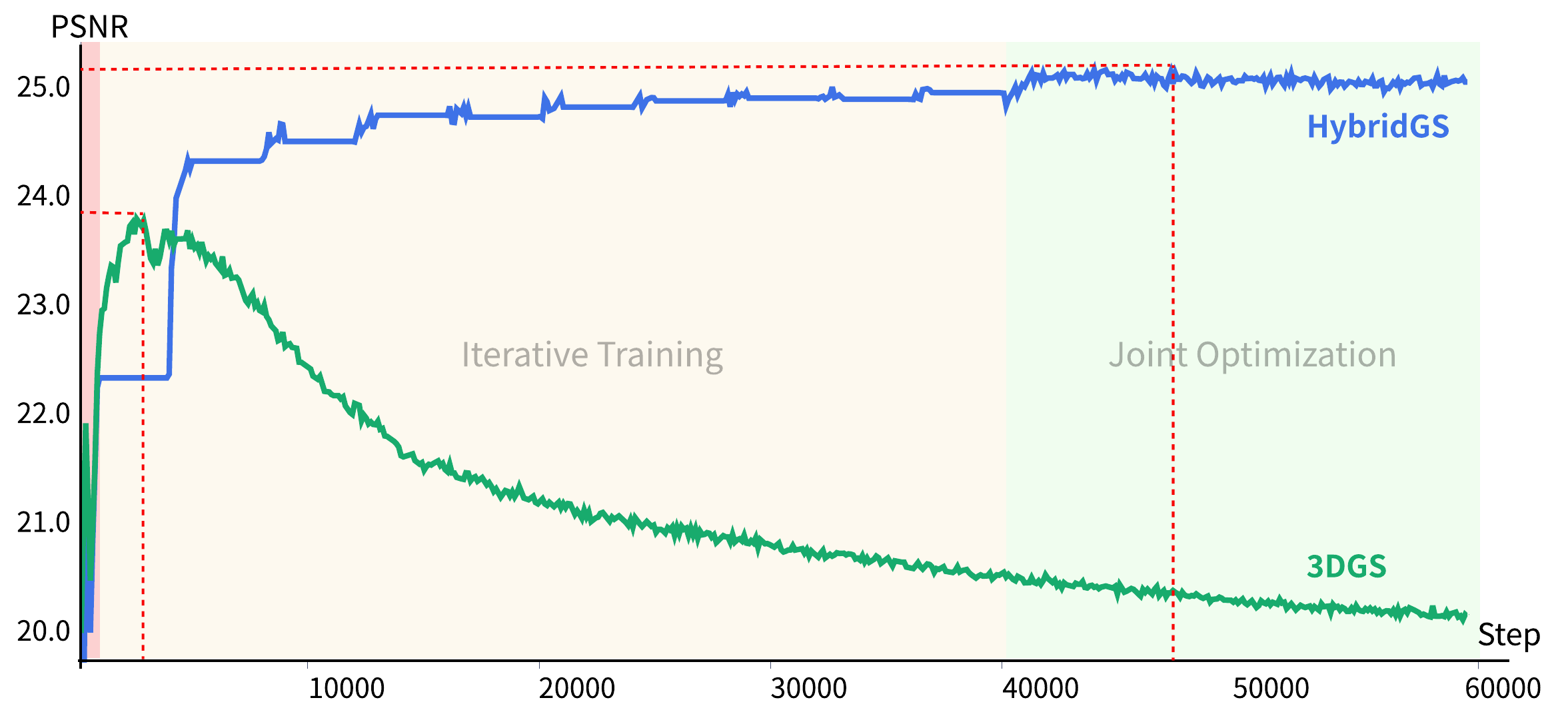}
  \caption{\textbf{The PSNR on testing set of Corner during training process.} Training steps for each stage of our HybridGS: \colorbox{myred}{Warm-up}: 0$\sim$1,010, \colorbox{myyellow}{Iterative Training}: 1,010$\sim$40,400, \colorbox{mygreen}{Joint Training}: 40,400$\sim$60,600. We use the same data as in Fig.~\ref{fig:training}. Note that, 3DGS reaches its optimum at step 2953, but as training continues, 3DGS tends to overfit transients in dynamic scenes, leading to gradual decline in performance. In comparison, our method is able to train steadily. This directly validates the statements in Sec.~\ref{subsec:2dgs} and \ref{subsec:training_process}.}
  \label{fig:psnr}
\end{figure}

To better demonstrate the changes during our training process, we select \texttt{IMG\_7195.JPG} of \textit{Corner} from NeRF On-the-go dataset as example, visualizing the statics and transients during different training stages and comparing them with vanilla 3DGS in Fig.~\ref{fig:training}. 
As training iterations increase, 3DGS tends to gradually integrate transient elements into the static components, rendering the residuals being almost incapable of capturing transient contents. In contrast, our HybridGS effectively distinguishes transients from statics over time, leading to consistent improvements
% in performance metrics as shown 
in Fig.~\ref{fig:psnr}.

\subsection{More Scenes}
In addition to providing metrics and results on the 6 commonly used scenes of NeRF On-the-go~\cite{Ren2024NeRF} dataset, we also present the visualization results on the remaining scenes as shown in Fig.~\ref{fig:on-the-go}. These complex scenes include some variations in lighting and shadows. We find that in addition to removing dynamic objects, our statics can also eliminate elements lacking specific semantics, such as shadows of pedestrians (in \textit{Drone} and \textit{Train Station}) and cars (in \textit{Arcdetriomphe} and \textit{Train}). 
This separation of non-semantic transients illustrates that our method is fundamentally a versatile, low-level and semantics-free scene decomposition approach, effectively highlighting its generality and robustness.

\subsection{More Datasets}
We apply our method on Photo Tourism~\cite{SSS:2006} dataset, which consists of unconstrained photo collections with photometric variations. As shown in Fig.~\ref{fig:photo}, we have some intriguing and reasonable observations. First, the statics generated using 3D Gaussians are rendered under an average light condition derived from the training images, similar to the diffuse lighting on an overcast day. Additionally, we discover that besides modeling dynamic objects, 2D Gaussians also capture photometric differences in our transients, since the illumination difference is indeed a per-image characteristic. This finding perfectly aligns with the perspective we presented in Sec.~\ref{subsec:2dgs} that transients can capture unique aspects of each image, broadening the scope for future research to further isolate photometric information from 2D Gaussians.

\begin{figure*}
  \centering
  \includegraphics[width=1.0\textwidth]{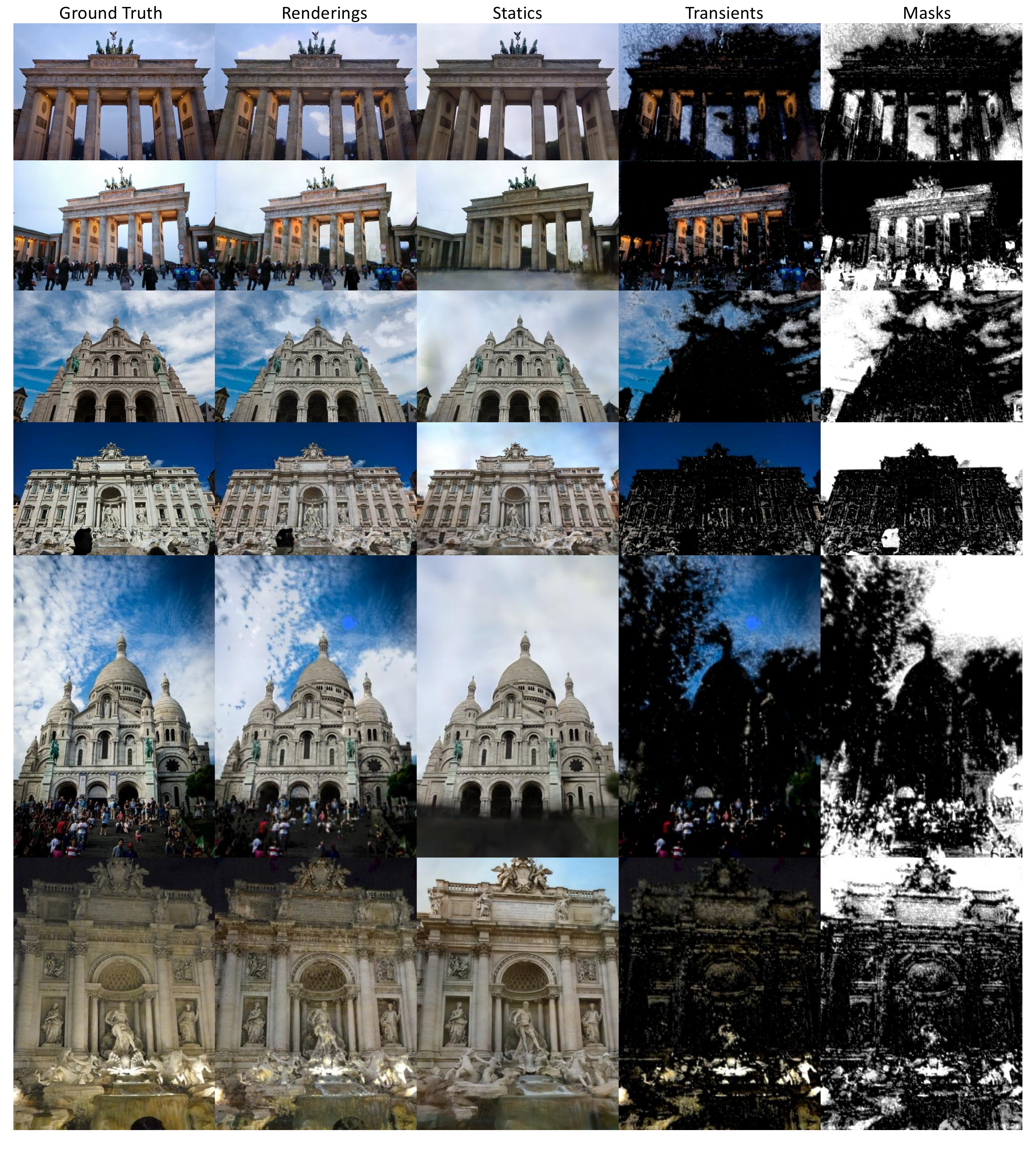}
  \caption{\textbf{Visualization on Photo Tourism~\cite{SSS:2006}.}}
  \label{fig:photo}
\end{figure*}

\end{document}